\documentclass[12pt]{article}

\usepackage{PRIMEarxiv}

\usepackage[utf8]{inputenc} 
\usepackage[T1]{fontenc}    
\usepackage{hyperref}       
\usepackage{url}            
\usepackage{booktabs}       
\usepackage{amsfonts}       
\usepackage{nicefrac}       
\usepackage{microtype}      
\usepackage{lipsum}
\usepackage{fancyhdr}       
\usepackage{graphicx}       
\usepackage{amssymb}
\usepackage[style=numeric-comp,sorting=none]{biblatex} 
\usepackage{amsmath}
\usepackage{multirow}
\graphicspath{{media/}}
\usepackage{subcaption}
\usepackage{algorithm}
\usepackage{algpseudocode} 

\title{PT: A Plain Transformer is Good 
Hospital Readmission Predictor}

\author{
  Zhenyi Fan$^\ast$ \\
  The University of Hong Kong \\
  \And
  Jiaqi Li$^\ast$ \\
  The University of Hong Kong \\
  \And
   Dongyu Luo$^\ast$ \\
  The University of Hong Kong \\
  \And
  Yuqi Yuan$^\ast$ \\
  The University of Hong Kong \\
}

\begin{document}
\maketitle
\footnotetext{$^\ast$ Equal contribution. Authors are listed in alphabetical order by surname.}

\begin{abstract}
Hospital readmission prediction is critical for clinical decision support, aiming to identify patients at risk of returning within 30 days post-discharge. High readmission rates often indicate inadequate treatment or post-discharge care, making effective prediction models essential for optimizing resources and improving patient outcomes. We propose PT, a Transformer-based model that integrates Electronic Health Records (EHR), medical images, and clinical notes to predict 30-day all-cause hospital readmissions. PT extracts features from raw data and uses specialized Transformer blocks tailored to the data's complexity. Enhanced with Random Forest for EHR feature selection and test-time ensemble techniques, PT achieves superior accuracy, scalability, and robustness. It performs well even when temporal information is missing. Our main contributions are: (1) \textbf{Simplicity}: A powerful and efficient baseline model outperforming existing ones in prediction accuracy; (2) \textbf{Scalability}: Flexible handling of various features from different modalities, achieving high performance with just clinical notes or EHR data; (3) \textbf{Robustness}: Strong predictive performance even with missing or unclear temporal data.

\end{abstract}

\section{Introduction}
Hospital readmission prediction is a critical component of clinical decision support, focusing on forecasting whether a patient will be readmitted to the same or another hospital within a specified period, typically ranging from 30 to 90 days post-discharge~\cite{wang2018predicting}. Readmissions often indicate that the initial hospitalization did not fully address the patient's health needs or that the patient's condition was inadequately managed after discharge. Accurate prediction models enable healthcare institutions to identify high-risk patients, optimize resource allocation, and implement personalized care interventions. Reducing unnecessary readmissions is essential for enhancing healthcare quality, lowering medical costs, and improving patient care, aligning with the core objectives of healthcare reform in numerous countries and regions~\cite{kansagara2011risk}. Effective prediction models and intervention strategies facilitate better long-term patient health management, decrease readmission rates, and enhance patients' quality of life~\cite{huang2019clinicalbert,kripalani2014reducing,merkow2015underlying}.

In recent years, the rapid advancement of deep learning has significantly influenced hospital readmission prediction research. Various deep learning methodologies, including Convolutional Neural Networks (CNN)~\cite{yang2018clinical,rajput2023based}, Recurrent Neural Networks (RNN)~\cite{lin2019analysis,reddy2018predicting}, Graph Neural Networks (GNN)~\cite{daneshva2022heterogeneous,golmaei2021deepnote}, and Long Short-Term Memory networks (LSTM)~\cite{reddy2018predicting,li2024predicting}, have been employed to extract features from Electronic Health Records (EHR) for predicting patient readmission risks. While EHRs provide structured data, clinical notes offer more comprehensive information. With advancements in deep learning, particularly in language models, research utilizing clinical notes has garnered significant attention~\cite{liu2019predicting,golmaei2021deepnote,brown2022information}. Notably, the ClinicalBERT~\cite{huang2019clinicalbert} model, pre-trained on clinical notes from the MIMIC-III dataset~\cite{johnson2016mimic}, has been fine-tuned to predict 30-day hospital readmissions with remarkable performance. Furthermore, the emergence of multimodal learning has led to the integration of various types of patient information for prediction. For example, Tang et al.~\cite{tang2022multimodal} extracted patient data from EHRs and clinical notes and employed Graph Neural Networks for readmission prediction. Compared to traditional baseline models such as XGBoost and LSTM, this multimodal approach significantly enhanced prediction accuracy.
\begin{figure}
    \centering
    \includegraphics[width=0.7\linewidth]{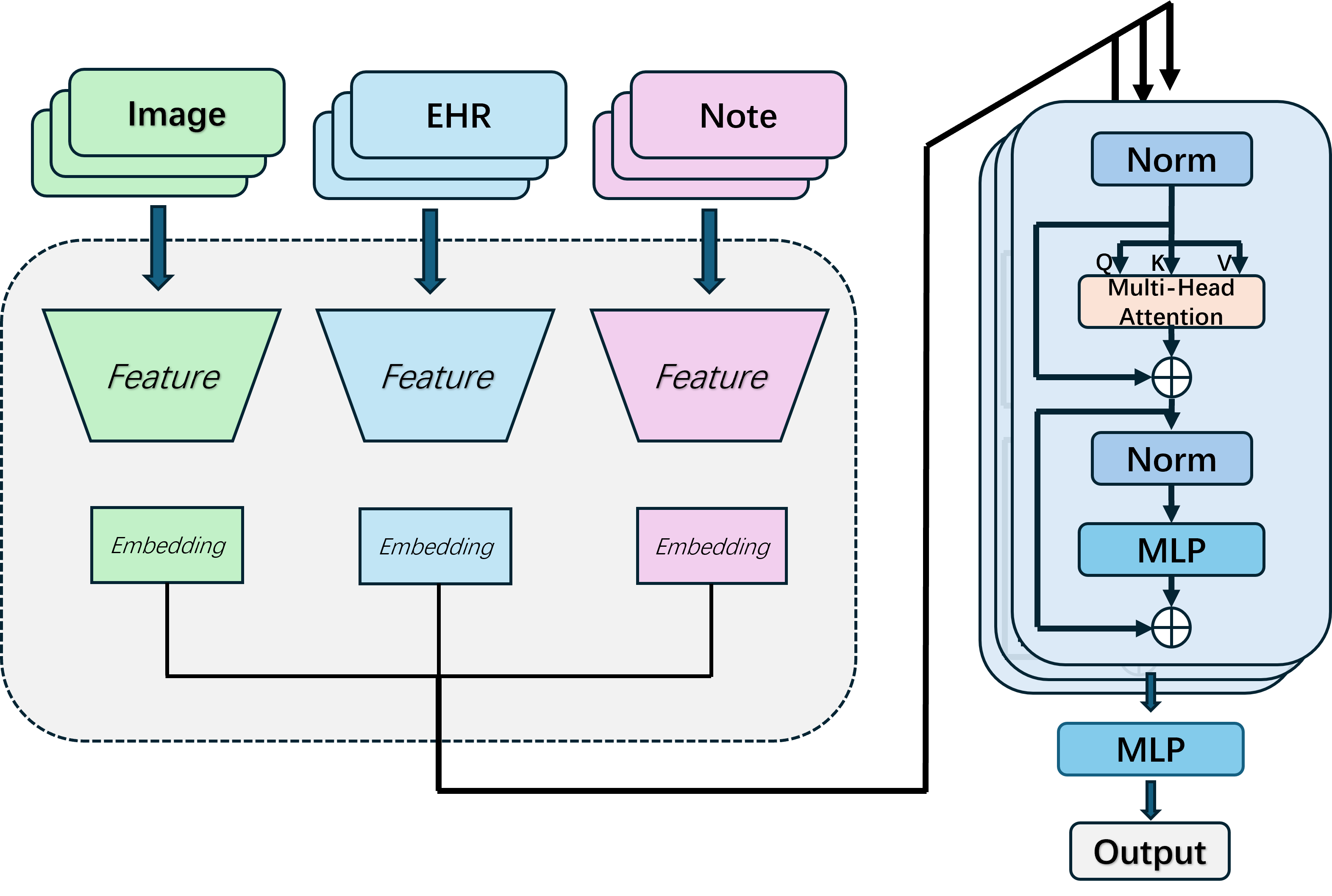}
    \caption{Overview of the Plain Transformer structure. Features from EHR modality follow previous preprocessing pipeline ~\cite{tang2022multimodal}.Image features for embedding are extracted with MoCo-CXR ~\cite{sowrirajan2021moco}. Clinical notes features are extracted and processed with TF-IDF text feature extractor ~\cite{qaiser2018text} ~\cite{ramos2003using}. Resulting features from the three modalities are fed into the transformer blocks for the final readmission prediction with each admission.}
    \label{fig:Figure 1: Plain Transformer}
\end{figure}
Currently, Transformer~\cite{vaswani2017attention} models have demonstrated remarkable performance in areas such as natural language processing~\cite{gillioz2020overview, wolf2020transformers}, large-scale language models~\cite{achiam2023gpt, bai2023qwen}, and image generation~\cite{dosovitskiy2020image, peebles2023scalable}, and their application in the medical field~\cite{valanarasu2021medical, hatamizadeh2022unetr} is gradually increasing. Due to the excellent performance of the Transformer architecture, it has achieved significant success in handling multimodal tasks~\cite{gabeur2020multi,nagrani2021attention,prakash2021multi,chen2022mm}. For example, Yan Miao and Lequan Yu proposed the MuST~\cite{miao2023must}, which leverages a Graph Transformer architecture to jointly predict hospital readmission based on EHR, medical images, and clinical notes. However, most multimodal research imposes strict requirements on the temporal features of input data. In practice, clinical data is often difficult to obtain, especially data with clearly labeled timestamps. Furthermore, medical images are harder to acquire compared to EHR and clinical notes, and generating simulated medical images presents additional challenges.

To address this issue, we propose a powerful baseline model, PT, based on the Transformer architecture. PT effectively integrates EHR, medical images, and clinical notes to predict 30-day all-cause hospital readmission. We first extract features from the raw inputs of EHR, medical images, and clinical notes. Then, based on the extracted feature dimensions and complexities, we construct different Transformer blocks. The outputs of these blocks are subsequently passed through a fully connected layer to generate the final prediction. Additionally, we introduce custom modules tailored to enhance the model's performance, such as using Random Forest for feature selection in EHR and employing test-time ensemble during final testing.

Our contributions can be summarized as follows:

\begin{enumerate}
    \item \textbf{Simplicity}: We present a powerful and efficient Transformer-based baseline model that outperforms other baseline models in terms of prediction accuracy.
    \item \textbf{Scalability}: The simplicity of the model's structure provides excellent scalability. By adjusting the number of Transformer blocks, the model can flexibly handle varying numbers and types of features from different modalities. It achieves outstanding performance even when using only clinical notes or only EHR data.
    \item \textbf{Robustness}: The proposed model demonstrates strong robustness, achieving excellent predictive performance even when temporal information in EHR, chest radiographs and clinical notes is unclear or missing.
\end{enumerate}

\section {Related works}
\paragraph{EHR and Notes} Electronic Health Record (EHR) is a digital system for storing and managing patient health information, which is fundamental in understanding patients' health conditions\cite{wang2020deep}. Nevertheless, traditional analytical approaches may be ineffective in processing these datasets due to their complex spatial and temporal structures\cite{article}. As vital unstructured healthcare records, Clinical notes also serve as a significant source of detailed patient information. The notes often contain valuable insights about symptoms, diagnoses, and treatment responses\cite{info:doi/10.2196/12239}, but the inherent complexity of medical terminology, inconsistent writing styles, and the abbreviations and acronyms make these text-rich health narratives hard to process and analyze\cite{computers10020024}. In recent years, significant progress has been made to address these challenges. Innovative deep learning frameworks, particularly those incorporating temporal convolution and attention mechanisms, have demonstrated effectiveness in managing the temporal complexities of structured EHR data\cite{XIE2022103980}. Simultaneously, the emergence of healthcare-specific language models has dramatically improved the ability to extract meaningful information from clinical notes\cite{biomedinformatics4020062}. However, a innegligible gap still remains in our ability to fully capture and utilize complex temporal relationships lying in these resources that are essential for accurate outcome prediction.

\paragraph{Multimodal-data learning} In modern healthcare systems, a multifaceted collection of patient information is usually utilized to store and analyze the health condition of the patients, including but not limited to physiological measurements, diagnostic test results, pharmaceutical records, radiological data, and physician documentation. The synthesis of these diverse data streams has demonstrated improvements in clinical outcome prediction and decision support systems\cite{acosta2022multimodal, fan2024research}. Nevertheless, the inherent disparity in data structures and formats poses challenges to effective multimodal integration\cite{fan2024research}. Traditional approaches to do multimodal learning mainly employed posterior fusion techniques, combining individual modality representations through simple concatenation operations or Kronecker product transformations\cite{HOANGTRONG2020105506}. Although these methods offer computational simplicity, they frequently fall short in capturing the nuanced relationships between different data modalities, resulting in incomplete learning. Based on these foundational studies, numerous models have been invented to improve the ability in multimodal data processing. For instance, Tang et al. \cite{tang2022multimodal} explored the use of Multimodal Spatiotemporal Graph Neural Networks, which is notable for its ability to handle multimodal data simultaneously while considering the temporal nature of the data, showing significant advantages in hospital readmission prediction.

\section{Methodology}
\label{sec:headings}
  Figure \ref{fig:Figure 1: Plain Transformer} illustrates our proposed plain transformer (PT) model for this task. In our approach, we explore various possible combinations between the three modality: electronic health record (EHR) data, chest radiography and clinical notes. Each modality is treated as sequential data, which, although lacking explicit timestamps, follows a clear temporal order. For each modality, relevant features are extracted through flexible, modular methods that can be easily adapted to different feature extraction techniques, depending on the specific characteristics of the data. Each extracted features are then processed by separate Transformer blocks ~\cite{vaswani2017attention} for each modality to capture the inherent structure within the sequences. Notably, the architecture of these blocks is similar across modalities, but they have different parameter sets, which are designed to adapted to the specific sequential characteristics of each modality. Finally, the outputs from the Transformer blocks are passed through a Multi-Layer Perceptron (MLP) to compute the probability of readmission.

\subsection{Feature extraction and processing}

\paragraph{EHR} The EHR modality encompasses a range of tabular data, including demographics, ICD-10 coded comorbidities, lab tests, and medication records. The raw EHR features are preprocessed following the procedure outlined in ~\cite{tang2022multimodal}. The dimensional structure of the EHR data for each patient admission is represented as $E \in \mathbb{R}^{n \times d}$, where $n$ denotes the total number of days the patient remained in the hospital following admission, and $d$ represents the number of distinct EHR features. To avoid over-fitting, we apply a Random Forest feature selector ~\cite{breiman2001random}, which identifies the $k$ most significant features from the original EHR data. Consequently, the final processed dimensionality for the EHR modality per admission is reduced to $E_{\text{selected}} \in \mathbb{R}^{n \times k}$, where the dimensionality of the feature set is reduced to $k$ after feature selection.


\paragraph{CXR} The chest radiography modality contains multiple original radiographical images stored in DICOM format. Feature extraction from these images is performed using MoCo-CXR ~\cite{sowrirajan2021moco}, a self-supervised, pretrained model based on DenseNet121. This model efficiently captures essential features from the radiographic images. The dimensional representation for this modality per admission is $CXR \in \mathbb{R}^{q \times 1024}$, where $q$ represents the number of radiographic images associated with that particular admission, and $1024$ represents the feature dimensions extracted from each image. When dealing with multiple images for a single admission, we maintain the original sequence as recorded in the CSV datasets to preserve the temporal information crucial for our analysis.

\paragraph{Clinical notes} The clinical notes modality consists of the medical notes associated with each patient's historical hospital admissions, organized by the sequence of notes. We explore traditional text feature extraction using the TF-IDF method ~\cite{qaiser2018text} ~\cite{ramos2003using}, which does not require any supplementary information and yields features with a dimension of 1024. The extracted text features fed into the model are structured as a matrix of dimensions $M \in \mathbb{R}^{m \times d}$, where $m$ represents the number of historical notes for each patient and $d$ denotes the feature dimension of the notes.

\subsection{Multimodal Integration via Transformer Architecture}

We propose a unified framework that integrates three distinct modalities: Electronic Health Record (EHR), Chest X-ray (CXR), and Clinical Notes. The EHR modality is represented as $\mathbf{X}_{\text{EHR}} \in \mathbb{R}^{n \times d}$, where $n$ is the number of hospital days and $d$ is the feature dimension; the CXR modality is represented as $\mathbf{X}_{\text{CXR}} \in \mathbb{R}^{q \times 1024}$, where $q$ is the number of images and $1024$ is the feature dimension extracted by the MoCo-CXR model~\cite{sowrirajan2021moco}; the Clinical Notes modality is represented as $\mathbf{X}_{\text{notes}} \in \mathbb{R}^{m \times d}$, where $m$ is the number of historical notes and $d$ is the feature dimension (1024 or 4096 depending on the extraction method).

Each modality is processed independently through its respective Transformer~\cite{vaswani2017attention} block to capture sequential relationships. For each modality, we define the following operations:
\[
\mathbf{H}_{\text{EHR}} = \text{Transformer}(\mathbf{X}_{\text{EHR}}, \theta_{\text{EHR}}), \quad
\mathbf{H}_{\text{CXR}} = \text{Transformer}(\mathbf{X}_{\text{CXR}}, \theta_{\text{CXR}}), \quad
\mathbf{H}_{\text{notes}} = \text{Transformer}(\mathbf{X}_{\text{notes}}, \theta_{\text{notes}}).
\]

The outputs of each modality's Transformer are passed through an Attention Pooling mechanism:
\[
\mathbf{H}_{\text{EHR}}^{\text{pool}} = \text{AttentionPooling}(\mathbf{H}_{\text{EHR}}), \quad
\mathbf{H}_{\text{CXR}}^{\text{pool}} = \text{AttentionPooling}(\mathbf{H}_{\text{CXR}}), \quad
\mathbf{H}_{\text{notes}}^{\text{pool}} = \text{AttentionPooling}(\mathbf{H}_{\text{notes}}).
\]

The pooled feature vectors are then concatenated into a unified feature vector:
\[
\mathbf{H}_{\text{concat}} = \text{concat}(\mathbf{H}_{\text{EHR}}^{\text{pool}}, \mathbf{H}_{\text{CXR}}^{\text{pool}}, \mathbf{H}_{\text{notes}}^{\text{pool}}).
\]

This concatenated feature vector $\mathbf{H}_{\text{concat}}$ is then passed through a Multi-Layer Perceptron (MLP) for final prediction:
\[
\hat{y} = \text{MLP}(\mathbf{H}_{\text{concat}}).
\]
where $\hat{y}$ is the predicted probability of patient readmission.

Each Transformer module uses positional encoding to retain sequence position information and maps the input data to the Transformer input space through embedding layers, enhancing sequential dependencies.

This framework processes each modality's data independently, effectively integrates features through Transformer and attention pooling mechanisms, and ultimately makes the readmission prediction through MLP.

\subsection{Label smoothing focal loss}

We integrate Label Smoothing and Focal Loss as training criterion~\cite{szegedy2016rethinking, lin2017focal}. Label Smoothing softens the target labels, avoiding overconfident predictions by the model and improving its ability to handle hard-to-classify samples. Specifically, for a target label \( t \in \{0, 1\} \), label smoothing is applied as follows:
\[
t' = (1 - \text{smooth}) \cdot t + \frac{\text{smooth}}{2}
\]
where \( \text{smooth} \) is the smoothing factor, and \( t' \) is the smoothed target.

Focal Loss introduces a focusing term that reduces the weight of easy-to-classify samples, enabling the model to focus more on hard-to-classify examples. In the readmission task, this helps improve prediction accuracy for minority class patients (e.g., those with low but critical readmission risk). The formula for Focal Loss is:
\[
\text{FL} = - \alpha (1 - p_t)^\gamma \log(p_t)
\]
where \( p_t = \sigma(z) \) is the predicted probability, \( \sigma(z) \) is the sigmoid function, and \( z \) is the model's output. \( \alpha \) is the weighting factor for class imbalance, and \( \gamma \) is the focusing parameter.

The final loss function, combining label smoothing and focal loss with a reduction type of mean, is given by:
\[
\mathcal{L}_{\text{mean}} = \frac{1}{N} \sum_{i=1}^N \alpha (1 - p_{t_i})^\gamma \cdot \left[ - t'_i \log(p_{t_i}) - (1 - t'_i) \log(1 - p_{t_i}) \right]
\]
where \( t'_i \) is the smoothed target for the \(i\)-th sample, \( p_{t_i} \) is the predicted probability for the \(i\)-th sample, and \( \alpha \) and \( \gamma \) are the hyperparameters for focal loss. A more detailed explanation is presented in the Algorithm~\ref{alg:loss} in Appendix~\ref{sec:label}.

By combining these two techniques, Label Smoothing Focal Loss effectively addresses class imbalance, reduces overconfident predictions, focuses on hard-to-classify samples, and improves the generalization ability of the model. This makes it particularly well-suited for the readmission prediction task, which often involves complex data and long-tailed distributions.

\subsection{Training \& testing phase strategies}
During the training of our model, we employ several strategies to enhance performance and optimize outcomes. For feature selection, we utilize a random forest feature selector for the EHR data, enabling us to identify critical features that contribute significantly to prediction accuracy. In order to improve convergence during training, we applied a cosine scheduler to modulate the learning rate effectively. Additionally, we introduce dynamic noise into the features aiming to potentially enhances the model's generalization capabilities. Finally, we employed a K-fold ensemble approach to generate robust prediction outcomes, further increasing the model's stability. Together, these strategies significantly improve the model's effectiveness in handling complex medical data. See the following for a detailed conceptual illustration of random forest feature selector, dynamic noise and K-fold ensemble.

\paragraph{Random Forest Feature Selection}
We employ a Random Forest feature selector \cite{breiman2001random} to identify the most informative features from the dataset. The feature selection process first begins by aggregating the feature representations for each patient in the training set. Specifically, we compute the mean feature values across all samples for each patient, resulting in a feature matrix \( X_{\text{train}} \in \mathbb{R}^{m \times d} \), where \( m \) represents the number of training samples and \( d \) denotes the total number of features.

The target labels corresponding to the training samples are stored in a vector \( y_{\text{train}} \in \mathbb{R}^{m} \).

Next, we initialize and train the Random Forest classifier with 100 trees, denoted as \( n_{\text{estimators}} = 100 \). The model is then trained on the feature matrix and the labels:
\[
\mathbf{rf.fit}(X_{\text{train}}, y_{\text{train}})
\]
After training, we extract the feature importance which indicates the significance of each feature in contributing to the model's predictions. The feature importance is given by the vector \( \mathbf{I} \in \mathbb{R}^{d} \), where each element corresponds to a feature's importance score.

To select the most relevant features, we sort the importances in descending order:
\[
\mathbf{J} = \text{argsort}(\mathbf{I}) \quad \text{(sorted indices)}
\]
Finally, we choose the top \( k \) features based on their importance scores:
\[
\mathbf{S} = \mathbf{J}_{1:k} \quad \text{(top $k$ features)}
\]
This selection process allows us to reduce the dimensionality of our dataset while retaining the most critical information, thus enhancing the model's efficiency and performance in subsequent analyses.

\paragraph{Dynamic Noise}

In our model training, we also introduce dynamic noise \cite{wikner2022stabilizing} to enhance feature robustness. We define a scheduling function \( \phi \) that determines the dynamic noise ratio for each epoch based on the original feature set \( \mathbf{F}_{\text{modality}} \in \mathbb{R}^{m \times d} \), where \( m \) denotes the number of patients and \( d \) indicates the number of distinct features. Thus, the noise ratio can be expressed as:
\[
r_{\text{epoch}} = \phi(\mathbf{F}_{\text{modality}})
\]
This scheduling function allows for flexibility in determining the noise levels, which can be adjusted dynamically to suit different training needs. For instance, one could implement a linear adjustment for the noise over the warm-up period, such that:
\[
r_{\text{epoch}} = 
\begin{cases} 
r_{\text{initial}} \left(1 - \frac{\text{epoch}}{W}\right) + r_{\text{final}} \left(\frac{\text{epoch}}{W}\right) & \text{if } \text{epoch} < W \\
r_{\text{final}} & \text{otherwise}
\end{cases}
\]
In this example, \( r_{\text{initial}} \) represents the starting noise ratio, while \( r_{\text{final}} \) denotes the ending noise ratio after the warm-up period. However, it is essential to note that these values are not fixed; the initial and final noise settings can be adjusted dynamically based on the specific requirements of the training process. This flexibility in the configuration of dynamic noise has the potential to enhance the generalization capabilities of our model, allowing it to adapt to various data distributions better.

To compute the noise standard deviations, we use:
\[
\mathbf{S}_{\text{noise}} = (\max({F_{\text{modality}}}) - \min({F_{\text{modality}}})) \cdot r_{\text{epoch}}
\]
Next, we generate noise \( N \in \mathbb{R}^{m \times d} \) that follows a Gaussian distribution:
\[
N \sim \mathcal{N}(0, \mathbf{S}_{\text{noise}})
\]
The final feature set after adding dynamic noise is denoted as:
\[
F_{\text{noisy}} \in \mathbb{R}^{m \times d}, \quad \text{where } F_{\text{noisy}} = F_{\text{modality}} + N
\]
\paragraph{K fold ensemble}

In order to further increase the robustness of prediction, we employ a K-fold ensemble approach \cite{anguita2012k} \cite{fushiki2011estimation} with $K=10$. The K-fold cross-validation process can be described as follows:

We first partition the dataset into \( K \) subsets (or folds), where \( K \) is specified as \( K = 10 \):
\[
D = \{D_1, D_2, \ldots, D_K\}
\]
For each fold \( i \) (where \( i = 1, 2, \ldots, K \)):
\begin{itemize}
    \item Use \( D_i \) as the validation set.
    \item Combine the remaining \( K - 1 \) folds for training:
    \[
    D_{\text{train}}^{(i)} = \bigcup_{j \neq i} D_j
    \]
\end{itemize}

After training on each combination, we aggregate the predictions from all \( K \) models to boost overall performance.

\section{Experiments}

\subsection{Dataset}
We evaluated the predictive capability of PT using three modalities of patient data for hospital readmission prediction~\cite{Wang2024STAT3612}:

\begin{itemize}
    \item \textbf{Electronic Health Record (EHR) Data (Tabular Data)}: We utilized Electronic Health Record (EHR) data from MIMIC-IV v1.0~\cite{johnson2023mimic,goldberger2000physiobank}. This dataset encompasses key patient information, including demographics (age, gender, and ethnicity), comorbidities recorded as ICD-10 codes, laboratory test results, and medications administered during hospitalizations.
    
    \item \textbf{Chest Radiographs (Imaging Data)}: Corresponding chest radiographs were obtained from MIMIC-CXR-JPG v2.0~\cite{johnson2019mimic,goldberger2000physiobank}. These images provide additional insights into the patients' conditions, complementing the structured data from the EHR.
    
    \item \textbf{Clinical Notes (Text Data)}: The de-identified free-text clinical notes for each patient were accessed from MIMIC-IV-Note \cite{johnson2023mimic,goldberger2000physiobank}. These notes, written by healthcare providers, offer rich, unstructured data that may contain important contextual information not captured in the structured EHR.
\end{itemize}

\textbf{Dataset Overview}: Our dataset consists of 13,763 hospital admissions, encompassing 82,465 chest radiographs from 11,041 unique patients. Among these, 2,379 admissions resulted in readmission within 30 days of discharge, including cases where patients passed away during their hospital stay.

\subsection{Implementations}
All experiments are conducted on a single NVIDIA GeForce A100 (40G) GPU, using the AdamW~\cite{loshchilov2017decoupled} optimizer and the CosineAnnealingLR~\cite{loshchilov2016sgdr} scheduler. The initial learning rate is set to 1e-3, with a minimum learning rate of 5e-4, and training runs for 100 epochs. The primary experiments use EHR and clinical notes as input data. Regarding the Transformer configuration, both transformers have 3 heads, with the EHR Transformer consisting of 2 layers and the Notes Transformer consisting of 3 layers. The evaluation metric used is the Area Under the Receiver Operating Characteristic (ROC) Curve (AUC), with detailed information provided in Appendix~\ref{sec:met}.

\subsection{Main result}

To ensure a fair evaluation of our model's performance, we select several classical non-Transformer architectures, such as LSTM~\cite{graves2012long} and GRU~\cite{chung2014empirical}, as baselines. We apply identical design configurations to these baseline models, including attention mechanisms and position encoding, to match those used in PT. This standardization ensures comparability in terms of parameter size and functionality, minimizing the impact of design differences. The results, shown in Table~\ref{table:performance_comparison}, demonstrate that, with comparable parameter counts and runtime, the Transformer model outperforms the baselines in terms of performance.

\begin{table}[htpb]
\centering
\caption{Comparison of model performance: AUC, running time, and parameters for LSTM, GRU, and PT (Ours).}
\begin{tabular}{l c c c}
\hline
\multirow{2}{*}{\textbf{Model}} & \multirow{2}{*}{\textbf{AUC}$\uparrow$} & \multirow{2}{*}{\textbf{Running time (s/Epoch)}} & \multirow{2}{*}{\textbf{Parameters (M)}} \\ 
& & & \\ \hline
LSTM ~\cite{graves2012long} & 0.878 & 3.0 & 0.55 \\ 
GRU ~\cite{chung2014empirical} & 0.880 & 2.8 & 0.44 \\ 
\textbf{PT (Ours)} & \textbf{0.896} & \textbf{3.3} & \textbf{0.49} \\ \hline
\end{tabular}
\label{table:performance_comparison}
\end{table}

\begin{table}[htbp]
\centering
\caption{Performance comparison of different modality combinations in terms of AUC and running time (s/Epoch).}
\begin{tabular}{ccc}
\toprule
\multirow{2}{*}{Modality} & \multirow{2}{*}{AUC$\uparrow$} & \multirow{2}{*}{Running time (s/Epoch)} \\
& & \\
\midrule
EHR & 0.761 & 2.4 \\
Notes & 0.832 & 1.4 \\
EHR+Notes & 0.896 & 3.3 \\
EHR+Note+Image & 0.881 & 12.5 \\
\bottomrule
\end{tabular}
\label{table:modal}
\end{table}

To evaluate the capability of PT in handling different modalities of data, we conduct experiments with various input data types. Additionally, due to the missing or unclear temporal information in our dataset, including EHR, chest radiographs, and clinical notes, our experiments further demonstrate the model's robustness in dealing with temporal data gaps. The specific experimental results are presented in Table~\ref{table:modal}.

\begin{table}[htbp]
\centering
\caption{Comparison of different configurations using Linear Noise Scheduler.}
\begin{tabular}{c c c}
\toprule
\textbf{Initial Noise Ratio} & \textbf{Final Noise Ratio} & \textbf{AUC} \\
\midrule
0.01 & 0.1 & 0.900 \\
0.001 & 0.1 & 0.892 \\
0.1 & 0.01 & 0.886 \\
0.1 & 0.001 & 0.885 \\
\bottomrule
\end{tabular}
\label{table:linear_noise_scheduler}
\end{table}

\begin{table}[htbp]
\centering
\caption{Comparison of different configurations using Sinusoidal Noise Scheduler (period = 40).}
\begin{tabular}{c c c}
\toprule
\textbf{Amplitude} & \textbf{Intercept} & \textbf{AUC} \\
\midrule
0.05 & 0 & 0.897 \\
0.1 & 0.01 & 0.897 \\
0.1 & 0 & 0.895 \\
\bottomrule
\end{tabular}
\label{table:sin_noise_scheduler}
\end{table}

In our experiments, we compare the performance of different configurations using both Linear and Sinusoidal noise schedulers. The results for the linear noise scheduler are presented in Table~\ref{table:linear_noise_scheduler}, where we evaluate various combinations of initial and final noise ratios. The best performance is achieved with an initial noise ratio of 0.01 and a final noise ratio of 0.1, yielding an AUC of 0.900.

We also experiment with a sinusoidal noise scheduler to evaluate its impact on model performance. The sinusoidal adjustment provides a different dynamic, potentially allowing for smoother transitions in noise levels during training. The noise ratio \( r_{\text{epoch}} \) in this case can be expressed as:

\[
r_{\text{epoch}} = \phi(\text{epoch}) = A \cdot \sin\left(\frac{2 \pi}{\text{period}} \cdot \text{epoch}\right) + \text{intercept}
\]

The results for the sinusoidal noise scheduler are summarized in Table~\ref{table:sin_noise_scheduler}. The AUC values for the Sinusoidal scheduler, based on a period of 40, show competitive performance with the highest AUC reaching 0.897 for an amplitude of 0.05 and intercept of 0. The findings suggest that different noise scheduling strategies can influence model performance, highlighting the importance of exploring various configurations to optimize generalization capabilities.

\begin{figure}[htbp]
    \centering
    \begin{subfigure}[b]{0.45\linewidth}
        \centering
        \includegraphics[width=\linewidth]{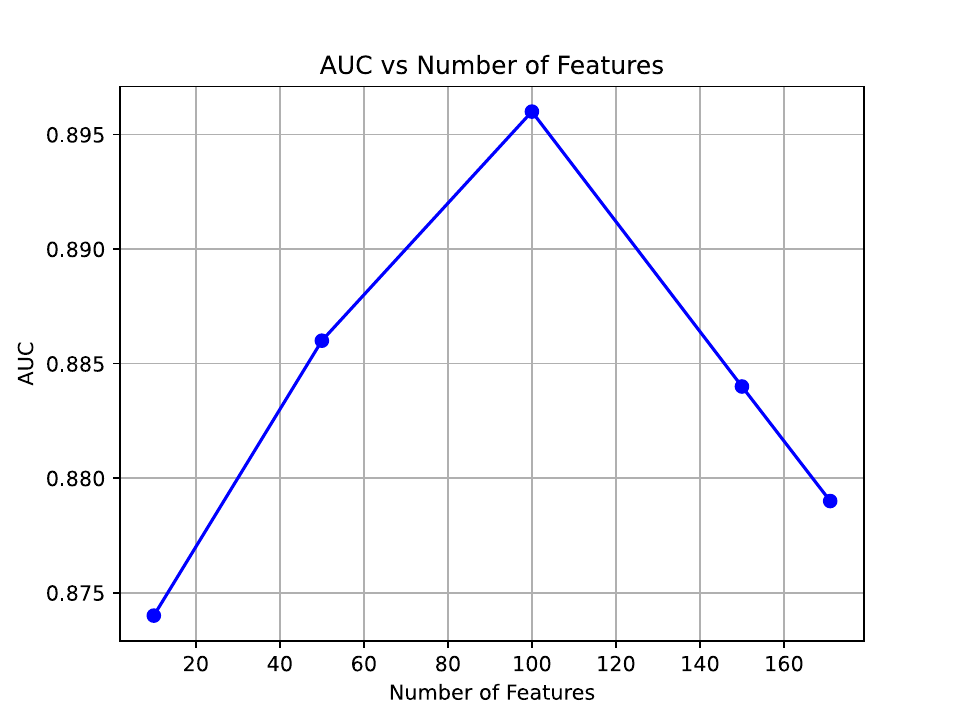}
        \caption{}
        \label{fig:num_fea_1}
    \end{subfigure}
    \hspace{0.1cm}
    \begin{subfigure}[b]{0.45\linewidth}
        \centering
        \includegraphics[width=\linewidth]{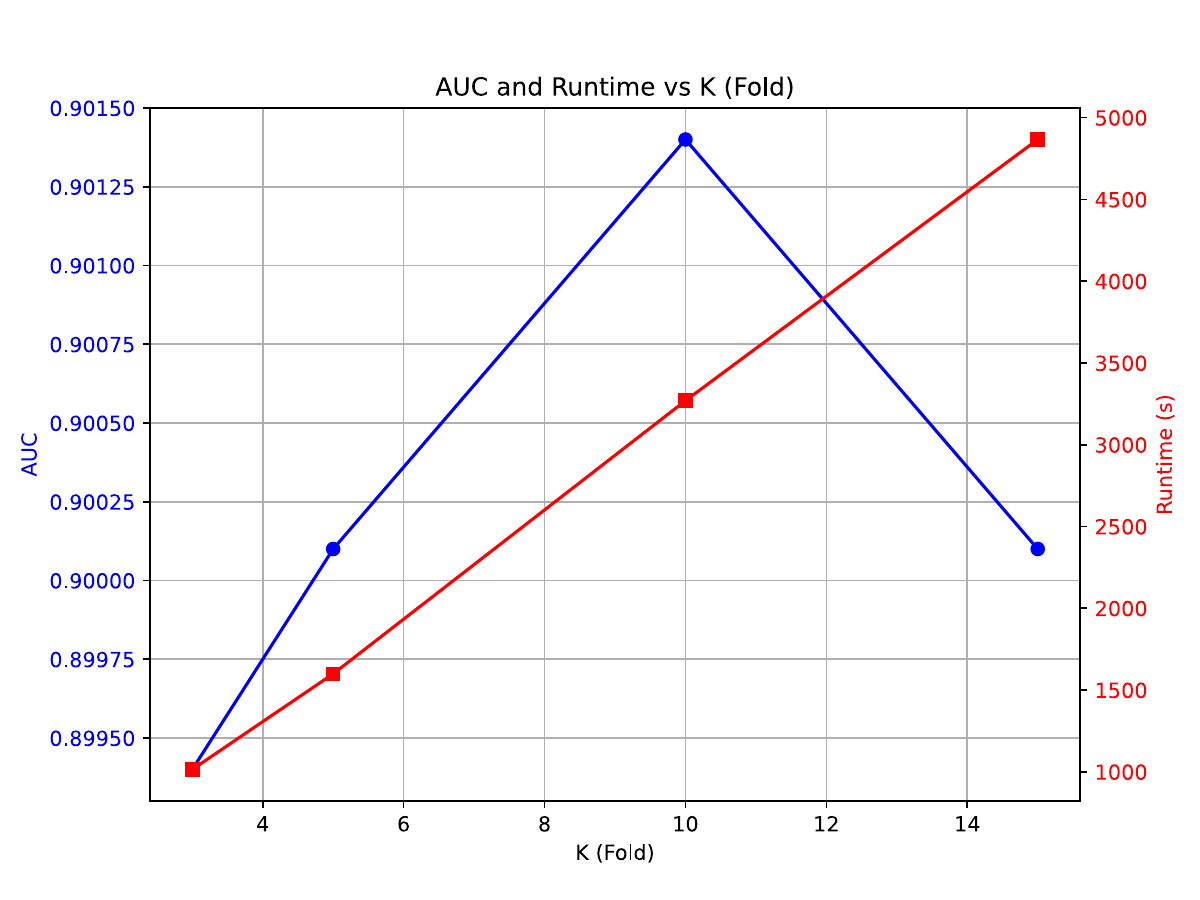}
        \caption{}
        \label{fig:num_fea_2}
    \end{subfigure}
    \caption{\textit{Left}: AUC performance of the random forest model with different numbers of selected EHR features; \textit{Right}: AUC performance of the model with different k values in K-fold cross-validation.}
    \label{fig:ablation_study}
\end{figure}

\subsection{Ablation study}
To further evaluate the performance of our model, we conduct a series of ablation studies. First, we analyze the impact of using different numbers of EHR features on model performance with Random Forest. As shown in Fig.~\ref{fig:num_fea_1}, the model performs best when the number of features is set to 100. To further investigate whether k-fold cross-validation can enhance the model's performance, we test different values of k. As shown in Fig.~\ref{fig:num_fea_2}, the model's performance significantly improved when k is set to 10. Depending on the task requirements, we can flexibly balance accuracy and inference time. In addition to this, we also compare the performance of Focal Loss and Binary Cross-Entropy(BCE) as training loss functions. As shown in Table~\ref{table:loss}, Focal Loss further improve the model’s performance. More details about the ablation experiments can be found in Appendix~\ref{sec:ab}


\begin{table}[htpb]
\centering
\caption{Comparison of loss strategies: AUC for Binary Cross Entropy (BCE) and Label smoothing focal loss.}
\resizebox{0.35\textwidth}{!}{
\begin{tabular}{cc}
\toprule
\multirow{2}{*}{Loss Strategy} & \multirow{2}{*}{AUC$\uparrow$} \\
& \\
\midrule
BCE & 0.891 \\
Label Smoothing Focal Loss & 0.896 \\
\bottomrule
\end{tabular}
}
\label{table:loss}
\end{table}

\section{Conclusion}
Our study highlights the advantages of combining multiple data types to harness their unique strengths, enabling the model to detect patterns that might remain hidden when relying on single data sources. By employing a transformer architecture with an attention mechanism specifically designed for processing multi-modal data and sequences, we achieved strong predictive performance, as demonstrated in our results. These findings provide actionable insights for improving patient care. For example, patients identified as high-risk can be enrolled in enhanced follow-up programs and receive personalized treatment plans. Beyond patient care, these predictions also offer valuable guidance for optimizing healthcare system resources. Overall, integrating diverse data sources with advanced modeling approaches shows great potential for enhancing predictive analytics in clinical practice.

\newpage
\printbibliography 
\newpage
\appendix
\section{Metric}\label{sec:met}
In this paper, we use AUC as the metric. The AUC is the area under the ROC curve, and can be calculated as:
\[
AUC = \int_{-\infty}^{\infty} TPR(FPR) \, dFPR
\]
Where:
\(TPR\) (True Positive Rate) is calculated as:
\[
TPR = \frac{TP}{TP + FN}
\]
\(FPR\) (False Positive Rate) is calculated as:
\[
FPR = \frac{FP}{FP + TN}
\]
Here, \(TP\) is the number of true positives, \(FP\) is the number of false positives, \(TN\) is the number of true negatives, and \(FN\) is the number of false negatives. The AUC value indicates the model's ability to distinguish between positive and negative samples, with values closer to 1 indicating better performance.

\section{Label Smoothing Focal Loss pseudo-code}\label{sec:label}
In this appendix, we provide the pseudo-code for the implementation of the Label Smoothing Focal Loss function, which is used in the context of predicting patient readmission. This loss function combines label smoothing and Focal loss to enhance the model's performance, particularly when dealing with imbalanced data in medical contexts. Label smoothing helps mitigate overconfidence in the model's predictions by distributing a small portion of the probability mass to incorrect classes, which is important in scenarios where predictions of patient readmission can be biased. Focal loss, on the other hand, focuses on hard-to-classify examples, such as patients with ambiguous or rare conditions, by down-weighting easy examples. The pseudo-code below outlines the steps involved in this loss function, including the computation of binary cross-entropy loss, the application of label smoothing to adjust the target labels, and the final loss computation based on the chosen reduction method.
\begin{algorithm}[H]
\caption{LabelSmoothingFocalLoss}
\label{alg:loss}
\begin{algorithmic}[1]
\State \textbf{Input:} 
\State \hspace{1em} $\mathbf{inputs}$: model logits (size $N$) \Comment{Model logits (predictions)}
\State \hspace{1em} $\mathbf{targets}$: ground truth labels (size $N$) \Comment{True labels}
\State \hspace{1em} $\alpha$: weight factor for Focal loss \Comment{Weight factor for Focal loss}
\State \hspace{1em} $\gamma$: focusing parameter for Focal loss \Comment{Focusing parameter for Focal loss}
\State \hspace{1em} $\text{smooth}$: label smoothing factor \Comment{Label smoothing factor}
\State \hspace{1em} $\text{reduction}$: reduction type ('mean', 'sum', 'none') \Comment{Reduction type for loss ('mean', 'sum', 'none')}

\State \textbf{Step 1: Label Smoothing}
\State \hspace{1em} \texttt{targets} $\gets$ \texttt{targets} $\times (1 - \text{smooth}) + 0.5 \times \text{smooth}$ \Comment{Apply label smoothing to targets}

\State \textbf{Step 2: Binary Cross Entropy Loss}
\State \hspace{1em} $\text{BCE\_loss} \gets \text{binary\_cross\_entropy\_with\_logits}(inputs, targets, \text{reduction='none'})$ \Comment{Compute binary cross-entropy loss}

\State \textbf{Step 3: Focal Loss}
\State \hspace{1em} $\mathbf{pt} \gets \exp(-\text{BCE\_loss})$ \Comment{Compute predicted probability}
\State \hspace{1em} $\text{F\_loss} \gets \alpha \times (1 - pt)^\gamma \times \text{BCE\_loss}$ \Comment{Apply Focal loss formula to focus on hard examples}

\State \textbf{Step 4: Apply Reduction}
\If {$\text{reduction} = \text{'mean'}$}
    \State \hspace{1em} return mean($\text{F\_loss}$) \Comment{Return the mean of the loss}
\ElsIf {$\text{reduction} = \text{'sum'}$}
    \State \hspace{1em} return sum($\text{F\_loss}$) \Comment{Return the sum of the loss}
\Else
    \State \hspace{1em} return $\text{F\_loss}$ \Comment{No reduction, return the loss as is}
\EndIf
\end{algorithmic}
\end{algorithm}

\section{Ablation study}\label{sec:ab}
Table~\ref{tab:fea} shows the AUC values corresponding to different features when using the Random Forest model, while Table~\ref{tab:kfold} presents the AUC performance of the model under different k values in k-fold cross-validation.

\begin{table}[htpb]
\centering
\caption{Features Number with Random Forest Selection}
\begin{tabular}{cc}
\toprule
\multirow{2}{*}{Number of Features} & \multirow{2}{*}{AUC} \\
& \\
\midrule
10 & 0.874 \\
50 & 0.886 \\
100 & 0.896 \\
150 & 0.884 \\
171 & 0.879 \\
\bottomrule
\end{tabular}
\label{tab:fea}
\end{table}

\begin{table}[htpb]
\centering
\caption{AUC and Runtime for Different K (Fold) Values}
\begin{tabular}{ccc}
\toprule
\multirow{2}{*}{K (Fold)} & \multirow{2}{*}{AUC} & \multirow{2}{*}{Runtime (s)} \\
& & \\
\hline
3 & 0.8994 & 1014 \\
5 & 0.9001 & 1599 \\
10 & 0.9014 & 3269 \\
15 & 0.9001 & 4866 \\
\bottomrule
\end{tabular}
\label{tab:kfold}
\end{table}

\end{document}